%% file: main.tex
\documentclass[10pt,twocolumn,letterpaper]{article}

\usepackage{cvpr}              

\usepackage{graphicx}
\usepackage{amsmath}
\usepackage{amssymb}
\usepackage{booktabs}
\usepackage{float}
\usepackage{caption}
\usepackage{colortbl}
\usepackage{xcolor}
\usepackage{float}
\usepackage[export]{adjustbox}
\usepackage[symbol]{footmisc}

\usepackage[pagebackref,breaklinks,colorlinks]{hyperref}

\usepackage[capitalize]{cleveref}
\crefname{section}{Sec.}{Secs.}
\Crefname{section}{Section}{Sections}
\Crefname{table}{Table}{Tables}
\crefname{table}{Tab.}{Tabs.}

\renewcommand{\thefootnote}{\arabic{footnote}}

\input{shortcuts}

\def\cvprPaperID{10894} 
\def\confName{CVPR}
\def\confYear{2024}

\begin{document}
	\title{ET3D: Efficient Text-to-3D Generation via Multi-View Distillation}
	
	\author{Yiming Chen$^{1,2}$\qquad Zhiqi Li$^{1,3}$\qquad Peidong Liu$^{1,\dag}$ \\$^{1}$Westlake University\qquad $^{2}$Tongji University \qquad $^{3}$Zhejiang University\\
		{\tt\small \{chenyiming, lizhiqi49, liupeidong\}@westlake.edu.cn}}
	
	\twocolumn[{
		\maketitle
		\vspace{-1em}
		\setlength\tabcolsep{0.2pt} 
		\centering
		\begin{tabular}{*{2}c}
			\includegraphics[width=0.5\linewidth]{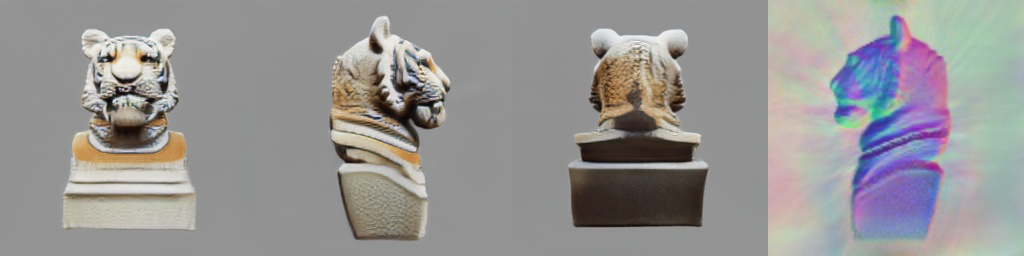} &
			\includegraphics[width=0.5\linewidth]{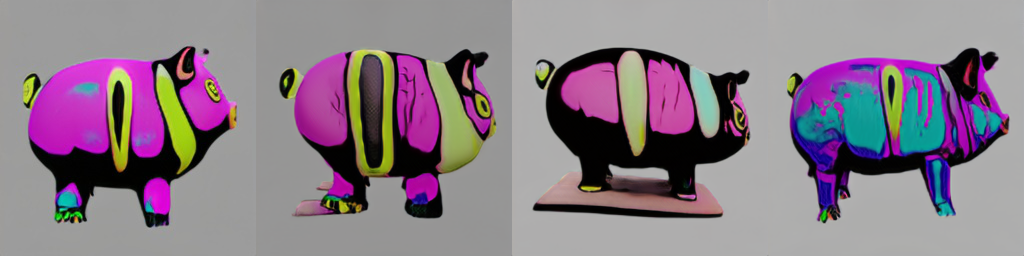} \\
			\small{(a) ``a stone bust of tiger"} & \small{(b) ``a pig, graffiti colors"} \\
			\includegraphics[width=0.5\linewidth]{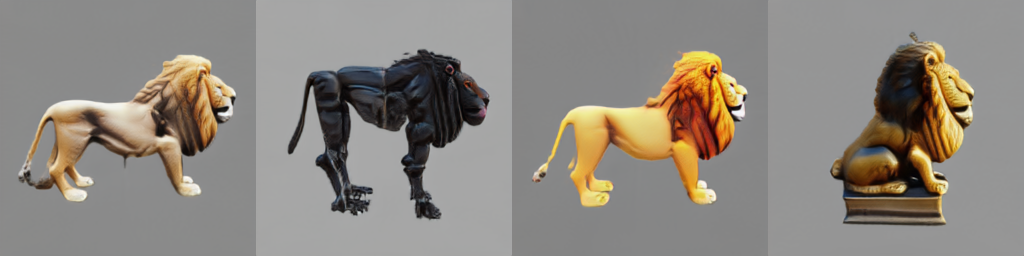} &
			\includegraphics[width=0.5\linewidth]{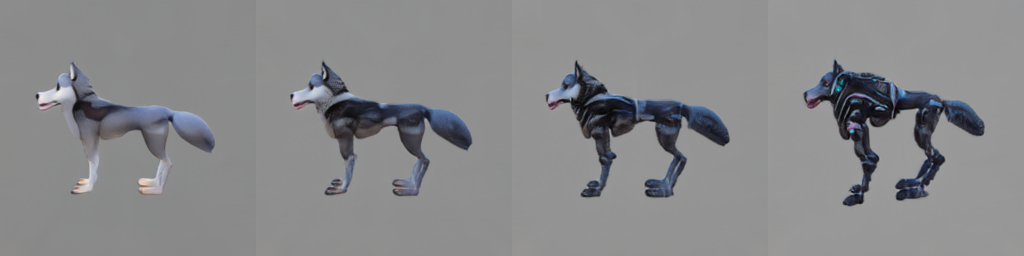} \\
			\small{(c) ``... photorealistic"\hspace{10pt}  ``... robot"\hspace{15pt} ``... cartoon"\hspace{10pt} ``... sculpture"} & 
			\small{(d) ``... disney style" $\rightarrow$ ``... robot"} \\
		\end{tabular}
		\vspace{-0.5em}
		\captionof{figure}{ET3D specializes in the efficient generation of 3D objects from text input, offering capabilities such as (a) producing multiview-consistent 3D objects conditioned on textual input, (b) generating diverse 3D objects with identical text and distinct latent inputs, (c) enabling style control in the output through text, and (d) facilitating smooth interpolations between prompts.}
		\vspace{1em}
	}]
        \let\thefootnote\relax\footnotetext{$^{\dag}$ Corresponding author.}

	\input{sec/0_abstract}    
	\input{sec/1_intro}

	\input{sec/2_related}
	\input{sec/3_method}

	\input{sec/4_exp}

	\input{sec/5_con}
	{
		\small
		\bibliographystyle{ieee_fullname}
		\bibliography{bibliograph_abbre,bibliography}
	}

\end{document}

%% file: shortcuts.tex
\newcommand{\bF}{\mathbf{F}} 

\newcommand{\bI}{\mathbf{I}}

\newcommand{\bK}{\mathbf{K}}

\newcommand{\bx}{\mathbf{x}}

\newcommand{\bz}{\mathbf{z}}

\newcommand{\bxi}{\boldsymbol{\xi}}

\newcommand{\nE}{\mathbb{E}}

\newcommand{\nR}{\mathbb{R}}

\newcommand{\cL}{\mathcal{L}}

\newcommand{\cN}{\mathcal{N}}

\newcommand{\figref}[1]{Fig.~\ref{#1}}

\newcommand{\tabnref}[1]{Table~\ref{#1}}

\makeatletter
\DeclareRobustCommand\onedot{\futurelet\@let@token\@onedot}
\def\@onedot{\ifx\@let@token.\else.\null\fi\xspace}
\def\eg{e.g\onedot} 
\def\ie{i.e\onedot} 
 
\def\etc{etc\onedot}

\def\etal{et~al\onedot}

\makeatother

\newcommand{\PAR}[1]{\vspace{0.1cm}\noindent{\bf #1} }

\newcommand{\norm}[1]{\left\lVert#1\right\rVert}

%% file: sec/0_abstract.tex
\begin{abstract}
Recent breakthroughs in text-to-image generation has shown encouraging results via large generative models. Due to the scarcity of 3D assets, it is hardly to transfer the success of text-to-image generation to that of text-to-3D generation. Existing text-to-3D generation methods usually adopt the paradigm of DreamFusion, which conducts per-asset optimization by distilling a pretrained text-to-image diffusion model. The generation speed usually ranges from several minutes to tens of minutes per 3D asset, which degrades the user experience and also imposes a burden to the service providers due to the high computational budget. 
In this work, we present an efficient text-to-3D generation method, which requires only around 8 $ms$ to generate a 3D asset given the text prompt on a consumer graphic card. The main insight is that we exploit the images generated by a large pre-trained text-to-image diffusion model, to supervise the training of a text conditioned 3D generative adversarial network. Once the network is trained, we are able to efficiently generate a 3D asset via a single forward pass. Our method requires no 3D training data and provides an alternative approach for efficient text-to-3D generation by distilling pre-trained image diffusion models. 

\end{abstract}

%% file: sec/1_intro.tex
\section{Introduction}

\begin{figure*}
    \setlength\tabcolsep{1pt} 
    \centering
    \begin{tabular}{*{3}c}
        \includegraphics[width=0.33\linewidth]{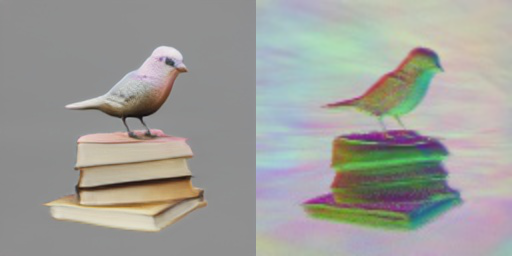} &
        \includegraphics[width=0.33\linewidth]{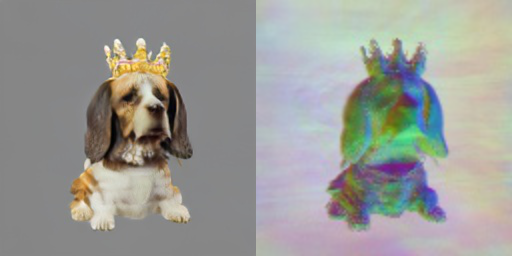} &
        \includegraphics[width=0.33\linewidth]{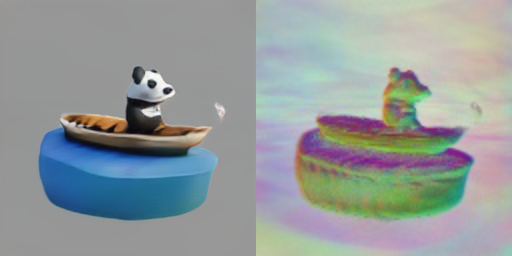} \\
        \small{a bird standing on a book} & \small{a cocker spaniel wearing a crown} & 
        \small{a panda rowing a boat} \\
        \includegraphics[width=0.33\linewidth]{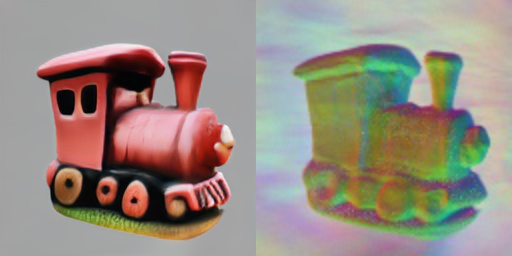} &
        \includegraphics[width=0.33\linewidth]{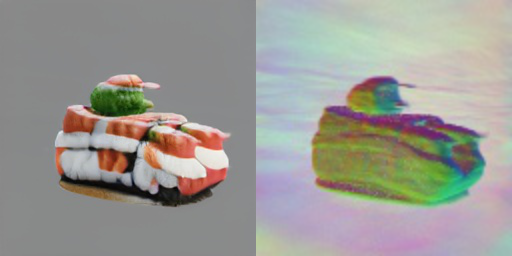} &
        \includegraphics[width=0.33\linewidth]{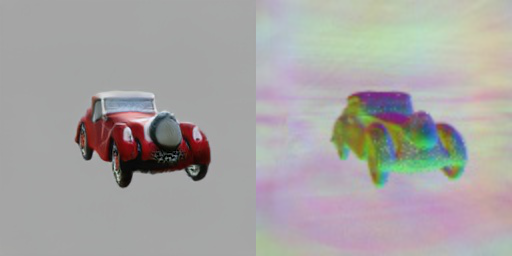} \\
        \small{a train engine made out of clay} & \small{a tank made of sushi} & 
        \small{a frazer nash super sport car} \\
    \end{tabular}
    \vspace{-0.5em}
    \caption{\textbf{Example 3D assets generated by ET3D.} Our network is able to generate text controlled 3D objects in only 8 $ms$.}
    \vspace{-1.5em}
\end{figure*}

\label{sec:intro}

Considerable advancements have been achieved in the realm of 2D image generation recently. The generation of high-fidelity images through input text prompts has become a straightforward process. However, the translation of this success from text-to-image generation to the text-to-3D domain faces challenges due to the limited availability of 3D training data.
To circumvent the need for training an extensive text-to-3D generative model from scratch, given the scarcity of 3D data, recent methods have capitalized on the favorable characteristics of diffusion models and differentiable 3D representations. These methods, rooted in score distillation sampling optimization (SDS), endeavor to extract 3D knowledge from a pre-trained, large text-to-image generative model, yielding impressive results. One notable example of such work is DreamFusion, which introduces a novel paradigm for 3D asset generation.

In light of the 2D-to-3D distillation approach, there has been a rapid evolution of techniques in the past year. Numerous studies have emerged, aiming to enhance the quality of generation through the implementation of multiple optimization stages. Although those methods are able to deliver impressive quality of the generated 3D objects, they usually require hours to finish the optimization process, which would degrade the user experience and also impose a burden to the service providers due to the requirement of more computational resources. 
To tackle the efficiency issue of existing text-to-3D generation methods, Lorraine \etal recently proposed ATT3D \cite{lorraine2023att3d}. The main insight is that they design a feed-forward mapping network, which maps the input text prompt to the parameters of a neural radiance field (NeRF). They can then render multi-view images from NeRF, and train the mapping network by SDS loss computed via a pre-trained 2D diffusion model. Once the network is trained, they are able to achieve efficient text-to-3D generation via a simple feed-forward pass. Due to characteristic of the used SDS loss, their method suffers from a lack of diversity and shared limitations with prior SDS-based works \cite{poole2022dreamfusion}. Another inspiring work is StyleAvatar3D \cite{zhang2023styleavatar3d}, they exploit a pre-trained ControlNet \cite{zhang2023controlnet} to generate multi-view images given a prior 3D head model and use those images to re-train EG3D \cite{chan2022eg3d} for 3D head generation. Since they require an existing 3D model for multi-view image generation, it is difficult for them to scale to general text-to-3D generation. 

Inspired by the recent development of large text-to-multi-view image generative models \cite{tang2023mvdiffusion,shi2023mvdream} and StyleAvatar3D \cite{zhang2023styleavatar3d}, we propose to train a text-to-3D generative model via multi-view distillation. The main insight is to exploit a pre-trained large image generative model as a teacher and distill multi-view knowledge to supervise the training of our text-to-3D model, \ie as a student network. In particular, we employ the pre-trained teacher network (\eg MVDream \cite{shi2023mvdream}) to generate multi-view images given a text prompt. We then train a text-conditioned generative adversarial network to generate a tri-plane represented 3D object, such that its rendered multi-view images follow the same distribution as that of the pre-trained text-to-multi-view model. Different from StyleAvatar3D \cite{zhang2023styleavatar3d}, our method does not require any prior 3D model and can scale to general text-to-3D generation task.   
%
Once our network is trained, we are able to generate a 3D object given a text prompt in only 8 ms on an NVIDIA RTX 4090 graphic card. It significantly accelerates the generation speed and reduces the computational expenses, to further democratize 3D content creation. 
%
%
In summary, our \textbf{contributions} are as follows:
\vspace{-0.4em}
\begin{itemize}
	\itemsep0em
	\item We propose a simple yet effective text conditioned 3D generative adversarial network;
	\item Our network can be trained by distilling multi-view knowledge from a pre-trained large text-to-multiview image generative model, without requiring SDS loss and any 3D dataset;
	\item Once our network is trained, it can generate a 3D asset given a text prompt, in only 8 ms on a consumer-grade graphic card. It significantly reduces the computational budget and provide the user with real-time experience;
	\item It demonstrates the possibility to train efficient general text-to-3D generative model by relying on pre-trained large text-to-multi-view image diffusion model;
	\item We would like to draw the attention of the community, that it would be a worthwhile direction to explore for efficient text-to-3D content generation, by exploiting pre-trained text-to-multi-view foundation models.
\end{itemize}

%% file: sec/2_related.tex
\section{Related Work}
\label{sec:related}
We review prior methods which are the most related to ours. We classify them into three categories: unconditional 3D generation, text conditioned 3D generation and 3D aware image synthesis.

\PAR{3D generative models.}
Unconditional 3D generation methods typically utilize existing 3D datasets to train generative models that employ various 3D representations. These representations commonly include volumetric  representation\cite{wu2016learning,brock20163dgan,gadelha20173d,li2019gan}, triangular mesh \cite{hamu2018siggraph,tan2018cvpr,gao2019tog,pavllo2021iccv,youwang2022eccv}, point cloud \cite{achlioptas2018icml,wu2019iccv,yang2019iccv,pumarola2020cvpr,Nichol2022arxiv}, and the more recent implicit neural representation \cite{park2019cvpr,mescheder2019cvpr,chen2019cvpr,schwarz2022neurips,chan2022eg3d,wang2023rodin}.
In the realm of 3D data, researchers have explored various generative modeling techniques that have demonstrated success in 2D image synthesis. These techniques encompass a range of methods, such as variational auto-encoders \cite{balashova2018VAE,wu2019ToG,tan2018cvpr,gao2019tog}, generative adversarial networks \cite{wu2016learning,li2019gan,phuoc2020neurips,pavllo2021iccv,achlioptas2018icml,chan2022eg3d}, flow-based methods  \cite{aliakbarian2022cvpr,yang2019iccv,pumarola2020cvpr,klokov2020eccv}, and the increasingly popular diffusion-based method \cite{luo2021cvpr,zhou2021iccv,zeng2022neurips,hui2022siggrapha,muller2023diffrf,chou2023diffsdf}. However, unlike image generative modeling, which benefits from a large abundance of training images, 3D generative methods often face a scarcity of sufficient 3D assets for training purposes. Typically, they are confined to category-specific datasets, such as shapeNet \cite{shapenet2015}. Although there has been a recent release of a million-scale 3D asset dataset by Objaverse \cite{deitke2023objaverse}, its size still pales in comparison to the vast amounts of 2D training data \cite{schuhmann2022laion} employed by modern generative models for image synthesis.
The limited availability of extensive training data poses a challenge for these 3D generative methods, as they struggle to generate arbitrary types of objects that can meet the diverse requirements of end consumers. In contrast to these methods that rely on copious amounts of 3D data, we propose an alternative approach that leverages a pre-trained large text-to-multi-view image generative model. By distilling multi-view knowledge, our proposed method aims to facilitate more generalized text-to-3D generation capabilities.

\PAR{Text conditioned 3D generation.}
Owing to the scarcity of 3D data, researchers have endeavored to extract knowledge for 3D generation by utilizing pre-trained large image models. Initially, efforts were made to employ a pre-trained CLIP model \cite{radford2021learning} to align the input text prompt with rendered images, aiming to supervise the process of 3D object generation \cite{sanghi2022clipforge, jain2022zero, mohammad2022clip}. However, the resulting 3D objects often exhibited a decreased level of realism, primarily due to the fact that CLIP could only provide high-level semantic guidance.
With the advancement of large text-to-image diffusion models \cite{saharia2022photorealistic}, a notable example being DreamFusion \cite{poole2022dreamfusion}, the potential to generate more realistic 3D objects through knowledge distillation has been demonstrated. Subsequent works have consistently pushed the boundaries to achieve the generation of photo-realistic 3D objects that closely correspond to the provided text prompts \cite{lin2023magic3d,wang2023SJC,chen2023fantasia3d,wang2023prolificdreamer,li2023sweetdreamer,tang2023dreamgaussian,zhu2023hifa,huang2023dreamtime,raj2023dreambooth3d}. These methods typically offer valuable insights by developing more sophisticated score distillation loss functions or by refining optimization strategies, to further enhance the quality of the generated objects.
Despite the success achieved by these methods in generating high-fidelity 3D shapes based on textual descriptions, they usually require hours to complete the text-to-3D shape generation process. It degrades the user experience and imposes additional economic burden to the service providers. Consequently, we propose to train an efficient text-to-3D generative model via multi-view distillation. Once our network is trained, we are able to generate 3D objects given text prompts in real-time on a consumer-grade graphic card.

\PAR{3D aware image synthesis.}
The exploration of extending 2D generative adversarial networks (GANs) \cite{goodfellow2014generative} to the realm of 3D has been extensively researched, primarily due to the advantage of not requiring a dedicated 3D dataset. A key concept behind this approach involves training a GAN capable of generating 3D representations based on 2D images.
Multiple forms of 3D representations have been investigated, including triangular mesh \cite{liao2020towards, szabo2019unsupervised}, volumetric representation \cite{henzler2019escaping, gadelha20173d, nguyen2019hologan, nguyen2020blockgan, wu2016learning, zhu2018visual}, and tri-plane \cite{chan2022efficient} \etc. Among these options, tri-plane stands out as an efficient choice due to its low memory consumption and fast image rendering, making it well-suited for GAN training. Moreover, there are alternative methods such as GANcraft \cite{hao2021gancraft}, which utilize sparse voxel grids for 3D scene generation, as well as fully implicit NeRF-based techniques that replace traditional generators with radiance fields \cite{Schwarz_Liao_Niemeyer_Geiger_2020, Chan_Monteiro_Kellnhofer_Wu_Wetzstein_2021, Zhou_Xie_Ni_Tian_2021, Niemeyer_Geiger_2021, Gu_Liu_Wang_Theobalt_2021}.
Although these approaches have demonstrated remarkable capabilities in generating high-quality 3D assets, they are often limited to class-specific tasks and lack the flexibility to enable control over the generation process through textual input. Consequently, we propose \textit{ET3D} for text-to-3D generation by distilling knowledge from a pre-trained large image diffusion model.

%% file: sec/3_method.tex
\section{Method}
%
\begin{figure*}
	\centering
	\includegraphics[width=0.99\linewidth]{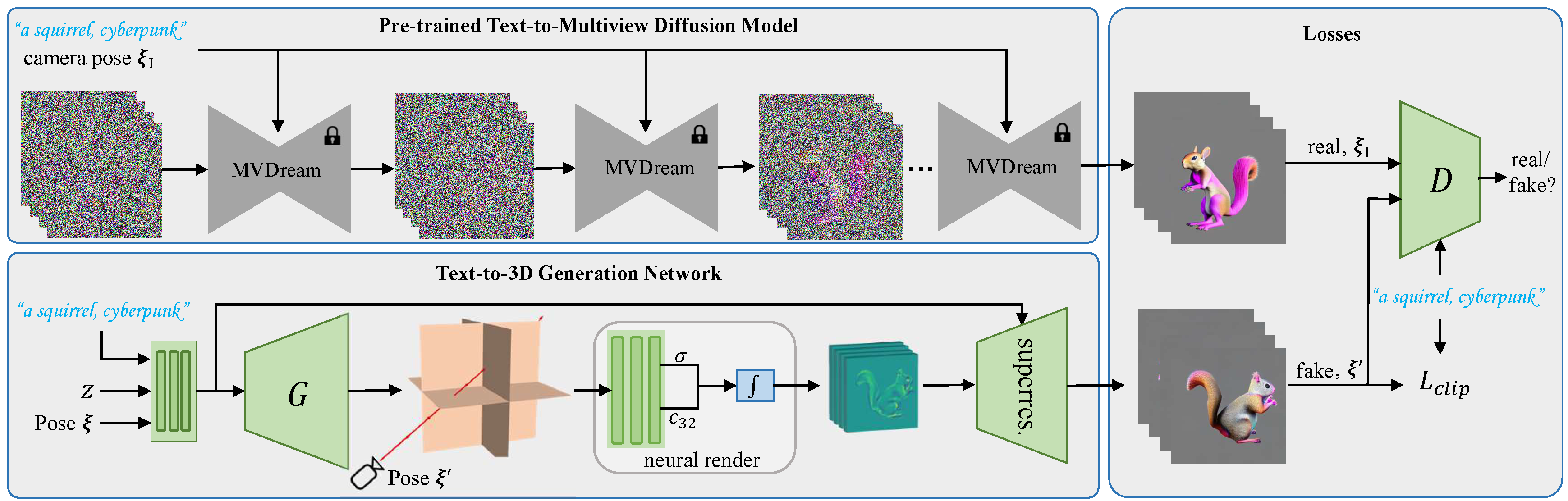}
	\caption{{\bf{The pipeline of ET3D.}} Our framework consists of a pretrained teacher model and a trainable student model. The student model learns to generate text controlled 3D asset via multi-view distillation of the teacher model. During training, both models receive the same text prompts. The teacher model is able to generate multi-view images. To train the student model, a discriminator network is used to supervise the generator network of the student model, to generate 3D content that has the same distribution as the teacher model in terms of rendered multi-view images. Once the student model is trained, it can generate 3D content in only 8 $ms$ on a consumer graphic card.} 
	\label{fig_pipeline}
	\vspace{-1em}
\end{figure*}

Our goal is to propose a new text-to-3D generation paradigm utilizing multi-view images synthesized by a large pre-trained image diffusion model. Although the text-to-multi-view diffusion model can generate impressive multi-view images, these images still lack pixel-wise consistency such that they can be used to reconstruct 3D assets directly. Instead of using Score Distillation Sampling (SDS) loss to align both data distributions, \ie which has shown to suffer from over-saturation, over-smoothing, low diversity and the multi-face Janus problem \cite{poole2022dreamfusion,wang2023prolificdreamer}, we propose to exploit Generative Adversarial Network (GAN) to learn the real data distribution.

Our method consists of two main parts as shown in Figure \ref{fig_pipeline}, \ie a teacher model and a student model. During training, both models accept the same textual input. The student model is then trained to generate a 3D asset which can render multi-view images, that follow the same distribution as that from the teacher model. The teacher model is a pre-trained text-to-multi-view image diffusion model. Since there is no prior text-to-3D GAN network available, we propose a simple yet effective network based upon EG3D \cite{chan2022eg3d}, due to its impressive performance in un-conditional 3D aware image synthesis from category-specific multi-view image dataset. We will detail each component as follows. 

\subsection{Text-to-multi-view image diffusion model.} Without loss of generality, we exploit a recently proposed text-to-multi-view image diffusion model, \ie MVDream \cite{shi2023mvdream}, as our teacher model. More advanced text-to-multi-view foundation models can also be used in future. MVDream is able to generate multi-view consistent images from a given text prompt. It achieves both the generalizability of 2D diffusion and the consistency of 3D data, by leveraging diffusion models pre-trained on large-scale web datasets and a multi-view dataset rendered from 3D assets. 

MVDream accepts a text prompt and four extrinsic camera parameters as input, it then generates four view-consistent images which satisfy the input text prompt each time. The current released pre-trained model enforces the four views to be $90^\circ$ separated for the longitude angle and share the same elevation angle, which ranges within $[0^\circ, 30^\circ]$. During the training of the student network, we sample multiple times for the same text prompt and the starting longitude angle is randomly selected within $[0^\circ, 360^\circ]$ each time. We note that the generated images are not always consistent between two samples (\ie 8 images in total) even the input text prompts are the same. However, we found that our student network is not affected and still can learn to generate 3D assets properly.

\subsection{Text-to-3D generative model.} Our student model is built upon EG3D \cite{chan2022eg3d}, a state-of-the-art 3D-aware image synthesis GAN network, which can learn from images only without requiring any 3D assets. As shown in \figref{fig_pipeline}, it consists of five key components: a mapping network, a tri-plane generator network, a neural renderer, a super-resolution module and a discriminator network. We will describe each component briefly as follows. More detailed network architecture can be found in our supplementary material.   

\PAR{Mapping network.} The mapping network takes a latent variable $z \in \nR^{512}$, camera parameters $P \in \nR^{25}$ and text embedding $T \in \nR^{768}$ as input, and map them into a 1280-dim feature vector. Both the latent variable and camera parameters are mapped into a 512-dim feature vector via EG3D's original MLP network. We then use a pre-trained CLIP model \cite{openclip2021} to encode the input text prompt into a 768-dim feature vector. Both feature vectors are then concatenated together to form the final 1280-dim feature vector for both the tri-plane generator network and the super-resolution module. 

\PAR{Tri-plane generator network.} By compromising the rendering efficiency and representation ability, we choose to use tri-plane \cite{chan2022eg3d} to represent the 3D object implicitly. The generator network takes the 1280-dim feature vector as input and outputs the tri-plane feature images, each with a dimension $\nR^{256\times256\times32}$.

\PAR{Neural renderer.} Given the generated tri-plane feature images and the sampled camera pose, we can render a 2D feature image $\bF \in \nR^{128\times128\times32}$ via volume rendering. In particular, we can shoot a ray from the camera center towards the sampled pixel. Discrete 3D points can be sampled along the ray. For each 3D point, we can project it into the tri-planes to obtain three feature vectors: \ie $\mathrm{F}_{XY}, \mathrm{F}_{XZ}$ and $\mathrm{F}_{YZ} \in \nR^{32}$. They are then concatenated together and input to a tri-plane decoder to obtain the point density $\sigma$ and a color feature vector $c \in \nR^{32}$. The pixel feature vector can then be computed via:
\begin{equation}
	\bF(\bx) = \sum_{i=1}^{n} T_i (1 - \mathrm{exp}(-\sigma_i \delta_i)) c_i,
\end{equation}
where $\bF(\bx) \in \nR^{32}$ is the rendered feature vector at pixel position $\bx$, $T_i$ is the transmittance and can be computed via $T_i = \mathrm{exp} (- \sum_{k=1}^{i-1} \sigma_k \delta_k)$, both $\sigma_i$ and $c_i$ are the predicted density and color feature vector of the sampled $i^{th}$ 3D point, and $\delta_i$ is the distance between two neighboring sampled points. 

\PAR{Super-resolution module.} To generate higher-resolution 3D assets, a super-resolution module is applied. It takes the rendered feature image $\bF \in \nR^{128\times128\times32}$ and the 1280-dim feature vector from mapping network as input, and predicts an image $\tilde{\bI} \in \nR^{256\times256\times3}$ as the final output image. 

\PAR{Discriminator network.} We modify the discriminator network of StyleGAN2 \cite{Karras2019stylegan2} to exploit text prompt embedding as additional condition to train the generator network. Same as the mapping network, we use a pre-trained CLIP model \cite{openclip2021} to encode the input text prompt, such that the discriminator can learn to differentiate images according to the provided text prompt. 

\subsection{Loss functions.} 
Both the generator network and discriminator network are trained in an adversarial manner. Given images $\bI$ from the pre-trained text-to-multi-view image diffusion model, with known camera parameters $\bxi_\bI$, $\bK$ for both the extrinsics and intrinsics, latent codes $\bz \in \cN(0, \boldsymbol{1})$ and the corresponding text prompts $t$, we train our model using a GAN objective. R1-regularization is applied to further stabilize the training \cite{mescheder2018ICML}: 
\small
\begin{align}
	&\cL(\theta,\phi) = \nE_{\bI \in p_D} (f(D_\phi (\bI, \bxi_\bI, t)) - \lambda \norm{\nabla D_{\phi}(\bI, \bxi_\bI,t)}^2 ) \\
	&+\nE_{\bz \in \cN(0, \boldsymbol{1}), \bxi, \bxi^{\prime} \in p_{\bxi}, t} [f(-D_{\phi} (G_{\theta}(\bz, \bxi, \bxi^{\prime}, \bK, t), \bxi^\prime, t))],
\end{align}
\normalsize
where $f(t) = -\mathrm{log}(1 + \mathrm{exp}(-t))$ and $\lambda$ controls the strength of the R1-regularizer. To better align the generated 3D asset with the textual description, we also apply a CLIP loss between the predicted image $\tilde{\bI}$ and the text prompt, which has shown to be effective in prior methods \cite{khalid2022clipmesh, wang2022clipnerf, Sauer2023ICML}. Both the generator and discriminator are then trained with alternating gradient descent combining the GAN objective with the CLIP loss:
\begin{align}
	&\min_{\theta} \max_{\phi} \cL(\theta,\phi) + \lambda_c \cL_{clip}(\theta), \\
	&\cL_{clip}(\theta) = \mathrm{arccos}^2(\mathrm{enc_i}(\tilde{\bI}), \mathrm{enc_t}(t)),
\label{formula_loss}
\end{align}
where $\lambda_c$ is a hyper-parameter, both $\mathrm{enc_i}$ and $\mathrm{enc_t}$ are the pre-trained CLIP image and text encoders.

%% file: sec/4_exp.tex
\section{Experiments}


\subsection{Implementation details}
Our framework offers the flexibility of being trained either online or offline with MVDream. We experimentally find that offline training still can deliver satisfying results, even the number of training samples would be much smaller than the online training. For efficiency consideration, we construct a substantial dataset with a wide variety of animals, objects etc, facilitating offline training in this experimental setup. The dataset comprises compositions of animals, objects and styles, totaling up to 5,000 different prompts and 800,000 generated images at a resolution of $256\times256$ pixels. We hold out 100 prompts during training and use them to evaluate the compositional generalization performance. Our method has no restriction to be scaled for even larger amount of text prompts. We use a learning rate $2.5\times10^{-3}$ for the generator training and $2\times10^{-3}$ to train the discriminator network. The batch size is 32. The network is trained with 8 NVIDIA A100 graphic card. All the evaluations are conducted on a single RTX 4090 graphic card.
We exploit the commonly used Frechet Inception Distance (FID) metric to evaluate the quality of rendered images from the generated 3D assets. CLIP score is used to evaluate the similarity between the input text prompt and the generated 3D asset.

\begin{figure*}[t]
	\setlength\tabcolsep{0.0pt}
	\centering
	\begin{tabular}{*{9}c}
		\small{a rabbit} & \small{a tiger}  & \small{a horse} & \small{a cat} & \small{a wolf} \\ 
		\includegraphics[width=0.1\linewidth]{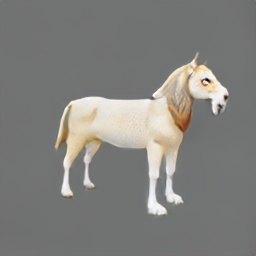} &
		\includegraphics[width=0.1\linewidth]{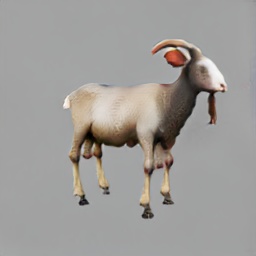} &
		\includegraphics[width=0.1\linewidth]{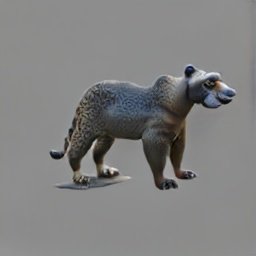} &
		\includegraphics[width=0.1\linewidth]{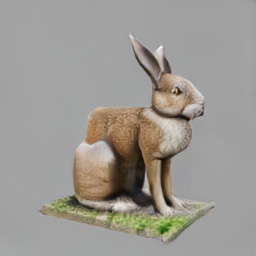} &
		\includegraphics[width=0.1\linewidth]{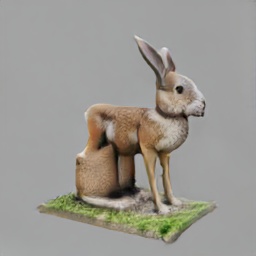} &
		\hspace{20pt} \includegraphics[width=0.1\linewidth]{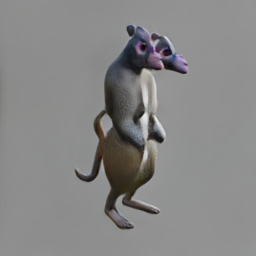} &
		\includegraphics[width=0.1\linewidth]{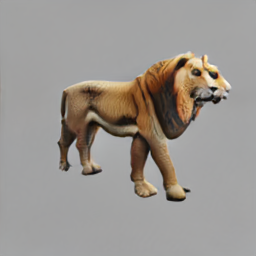} &
		\includegraphics[width=0.1\linewidth]{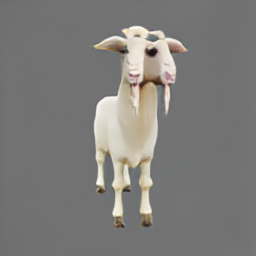} &
		\includegraphics[width=0.1\linewidth]{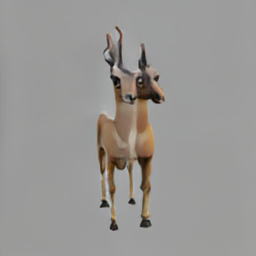} \\

		\includegraphics[width=0.1\linewidth]{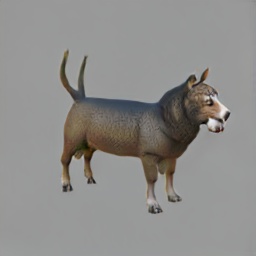} &
		\includegraphics[width=0.1\linewidth]{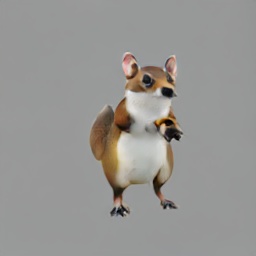} &
		\includegraphics[width=0.1\linewidth]{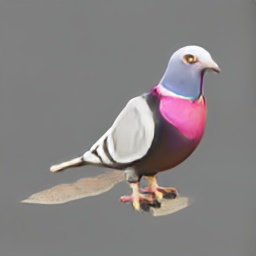} &
		\includegraphics[width=0.1\linewidth]{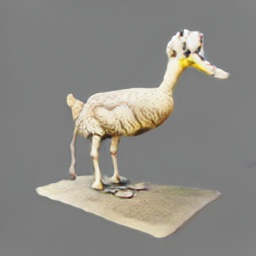} &
		\includegraphics[width=0.1\linewidth]{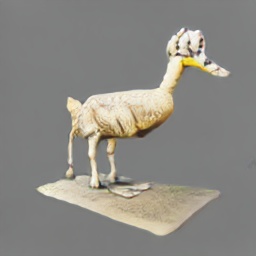} &
		\hspace{20pt} \includegraphics[width=0.1\linewidth]{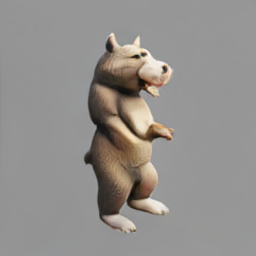} &
		\includegraphics[width=0.1\linewidth]{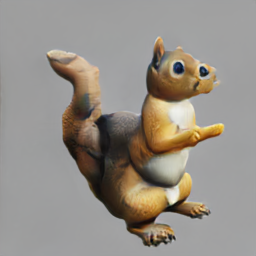} &
		\includegraphics[width=0.1\linewidth]{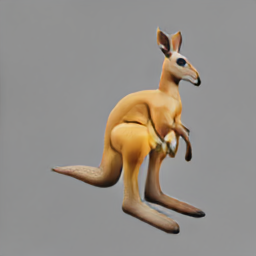} &
		\includegraphics[width=0.1\linewidth]{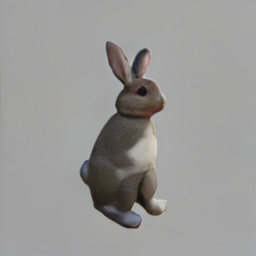} \\
		\multicolumn{5}{c}{w/o D text cond} & \multicolumn{4}{c}{EG3D}
	\end{tabular}
	\vspace{-0.5em}
	\caption{ {\textbf{Ablation study on the text condition.}} It demonstrates that the generator fails to learn proper text control if there is no text condition to be applied on the discriminator network. EG3D struggles to learn from such a large dataset containing many different types of objects. It results in 3D objects with multiple heads.}
	\label{fig_ablation}
	\vspace{-1.5em}
\end{figure*}

\begin{table}[t]
	\setlength\tabcolsep{5pt}
	\centering
	\begin{tabular}{c|ccc}
		\specialrule{0.1em}{1pt}{1pt}
		& FID$\downarrow$ & ViT-L/14$\uparrow$ & ViT-bigG-14$\uparrow$ \\ 
		\specialrule{0.05em}{1pt}{1pt}
		Ours & 7.4 & 0.305 & 0.45 \\
		w/o clip-loss & 7.7 & 0.317 & 0.452 \\
		w/o D text cond & 12.1 & 0.188 & 0.322 \\
		EG3D & 11.3 & $\times$ & $\times$ \\
		\specialrule{0.1em}{1pt}{1pt}
	\end{tabular}
	\vspace{-0.5em}
	\caption{{\bf{Ablation study.}} We study the effects of the CLIP loss and the text condition on the discriminator. The results demonstrate that the generator network struggles to learn text controlled generation without applying the text condition to the discriminator network. Due to the text conditioned discriminator already has strong capability to supervise text consistent 3D asset generation, the effect of CLIP loss in text control is marginal. However, it results in better FID score. We thus still adopt it for our training.}
	\label{table_ablation}
	\vspace{-1em}
\end{table}

\subsection{Ablation study} 
We conduct experiments to study the effects of the CLIP loss and conditional text input for the discriminator network. The experimental results are presented in \tabnref{table_ablation} and \figref{fig_ablation}. 
It demonstrates that the textual condition on the discriminator network improves both the quality and text coherence of the generated 3D assets. The qualitative results shown in \figref{fig_ablation} also demonstrate that the generated 3D assets fail to satisfy the input textual description if we do not apply the text condition on the discriminator. Unexpectedly, the usage of CLIP loss does not improve the similarity between the text description and generated 3D asset. The reason might be the text condition of the discriminator network can already provide sufficient supervision on text control. However, we find the FID metric is improved from 7.7 to 7.4 when the CLIP loss is applied. Therefore, we still keep the CLIP loss during training to obtain a better image quality. 

We also compare against EG3D \cite{chan2022eg3d}, which does not support text controlled 3D generation. The experimental results demonstrate that the EG3D struggles to learn unconditional 3D generation from dataset with many different categories. It demonstrates that the additional text condition helps the network cluster the data distribution and ease the learning of the generative network. 

\begin{table}[t]
	\setlength\tabcolsep{2pt}
	\begin{tabular}{c|ccc}
		\specialrule{0.1em}{1pt}{1pt}
		& ViT-L/14$\uparrow$ & ViT-bigG-14$\uparrow$ & Time(s)$\downarrow$ \\
		\specialrule{0.05em}{1pt}{1pt}
		DreamFusion-IF & 0.297 & 0.417 & 1800\\
		ProlificDreamer & 0.334 & 0.447 & 3600\\
		Shap-E & 0.265 & 0.347 & 5 \\
		Ours & 0.322 & 0.427 & 0.008 \\
		\specialrule{0.1em}{1pt}{1pt}
	\end{tabular}
	\vspace{-0.5em}
	\caption{{\bf{Quantitative Comparison.}} We compare ET3D against several prior state-of-the-art methods, including DreamFusion, ProlificDreamer and Shap-E. It demonstrates that ET3D is able to generate similar 3D content in terms of the CLIP score. However, its generation speed is greatly improved compared to those methods, \eg 225000 times faster than DreamFusion.}
	\label{tab_quant}
	\vspace{-1.5em}
\end{table}

\subsection{Quantitative comparisons}
We compare ET3D against prior state-of-the-art methods for quantitative evaluations. 
We exploit two state-of-the-art SDS optimization based methods, \ie DreamFusion \cite{poole2022dreamfusion} and ProlificDreamer \cite{wang2023prolificdreamer}. We use a rendering image resolution at $64\times64$ pixels for the optimization of ProlificDreamer. We also compare against Shap-E \cite{jun2023shape}, which is pre-trained with text-labeled 3D data. Another similar method is ATT3D \cite{lorraine2023att3d}. However, we cannot compare against it since they do not release their implementations to the general public. We exploit the CLIP score and time consumption as metrics for the evaluation. The experimental results are presented in \tabnref{tab_quant}. The metrics are computed over 400 different objects.
It demonstrates that our method is able to achieve similar or even better text similarity score, compared to DreamFusion, ProlificDreamer and Shap-E. On the other hand, the time required to generate a 3D asset by our method is only around 8 ms, which is 225000 times faster than DreamFusion and 450000 times faster than ProlificDreamer. The evaluations are conducted on a consumer graphic card, \ie NVIDIA RTX 4090.    

\begin{figure*}
	\setlength\tabcolsep{0.25pt} 
	\centering
	\begin{tabular}{*{4}c}
		\small{DreamFusion-IF} & \small{ProlificDreamer} & \small{Shap-E} & \small{ours}\\
		\includegraphics[width=0.25\linewidth]{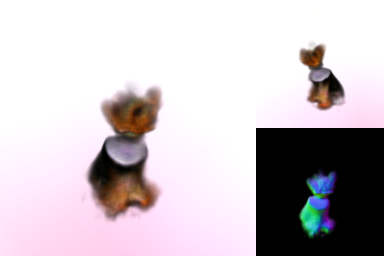} & \includegraphics[width=0.25\linewidth]{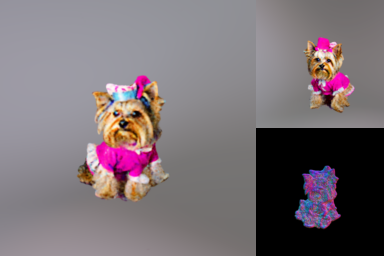} & 
		\includegraphics[width=0.25\linewidth]{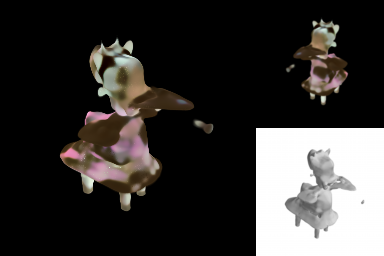} & \includegraphics[width=0.25\linewidth]{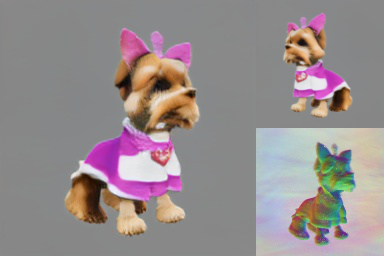} \\
	\end{tabular}
	\centering
	\small{``a yorkie dog dressed as a maid"} \\
	\vspace{0.3em}
	
	\begin{tabular}{*{4}c}
		\includegraphics[width=0.25\linewidth]{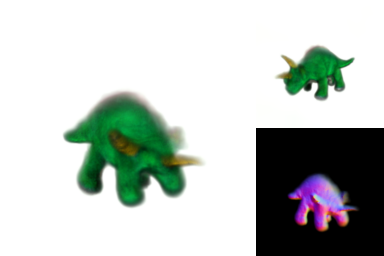} & 
		\includegraphics[width=0.25\linewidth]{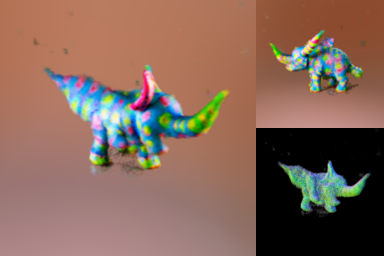} & 
		\includegraphics[width=0.25\linewidth]{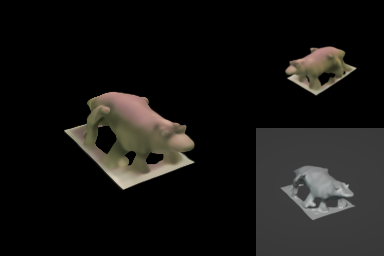} & \includegraphics[width=0.25\linewidth]{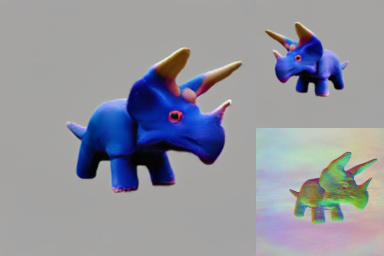} \\
	\end{tabular}
	\centering
	\small{``a plush triceratops toy, studio lighting, high resolution"} \\
	\vspace{0.3em}
	
	\begin{tabular}{*{4}c}
		\includegraphics[width=0.25\linewidth]{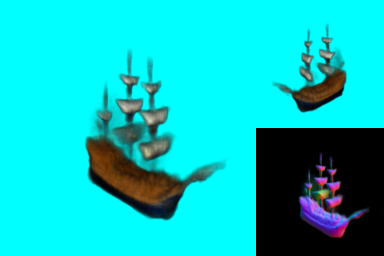} &
		\includegraphics[width=0.25\linewidth]{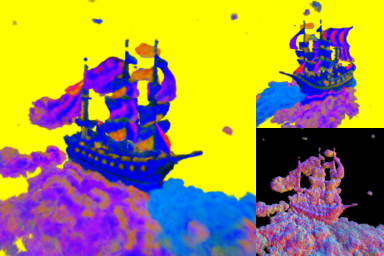} & 
		\includegraphics[width=0.25\linewidth]{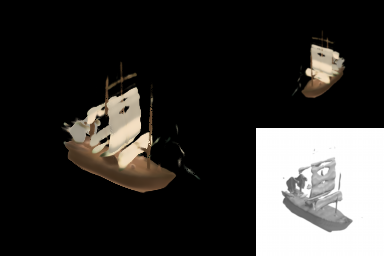} & \includegraphics[width=0.25\linewidth]{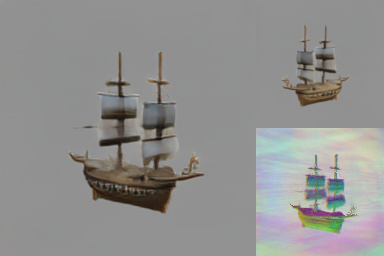} \\
	\end{tabular}
	\centering
	\small{``a spanish galleon"} \\
	\vspace{0.3em}
	
	\begin{tabular}{*{4}c}    
		\includegraphics[width=0.25\linewidth]{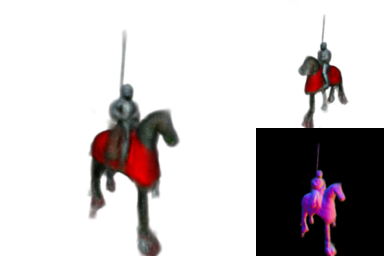} &
		\includegraphics[width=0.25\linewidth]{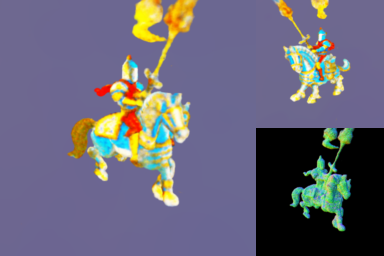} &
		\includegraphics[width=0.25\linewidth]{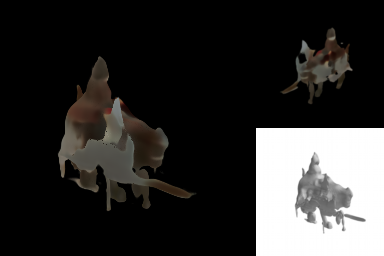} & \includegraphics[width=0.25\linewidth]{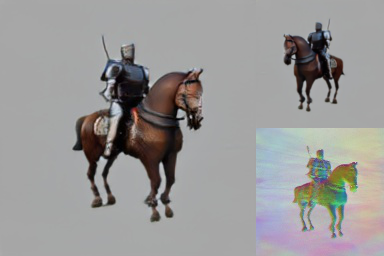} \\
	\end{tabular}
	\centering
	\small{``a knight holding a lance and sitting on an armored horse"} \\
	
	\vspace{-0.5em}
	\caption{ {\textbf{Qualitative comparisons.}} It demonstrates that DreamFusion tends to generate blurry 3D asset due to the SDS loss. While ProlificDreamer generates impressive 3D objects, it still fails on several text prompts. Our method generates view consistent 3D asset in good quality.}
	\label{fig_quality}
	\vspace{-0.7em}
\end{figure*}

\begin{figure*}
	\setlength\tabcolsep{0.0pt} 
	\centering
	\begin{tabular}{*{11}c}
		\includegraphics[width=0.09\linewidth]{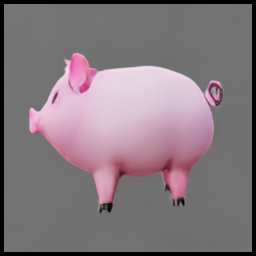} &
		\includegraphics[width=0.09\linewidth]{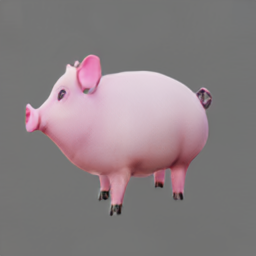} &
		\includegraphics[width=0.09\linewidth]{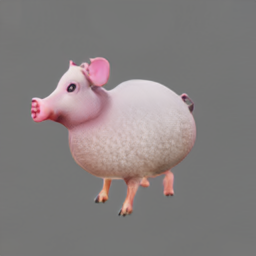} &
		\includegraphics[width=0.09\linewidth]{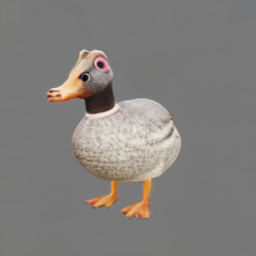} &
		\includegraphics[width=0.09\linewidth]{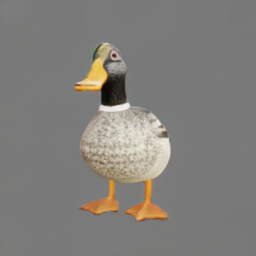} &
		\includegraphics[width=0.09\linewidth]{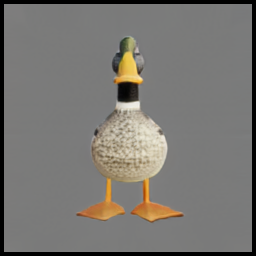} &
		\includegraphics[width=0.09\linewidth]{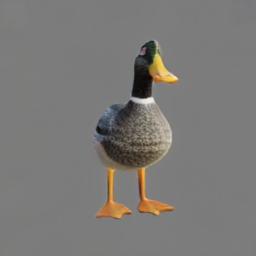} &
		\includegraphics[width=0.09\linewidth]{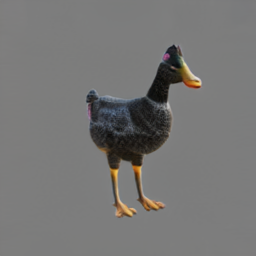} &
		\includegraphics[width=0.09\linewidth]{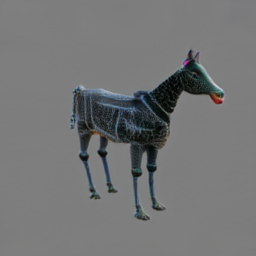} &
		\includegraphics[width=0.09\linewidth]{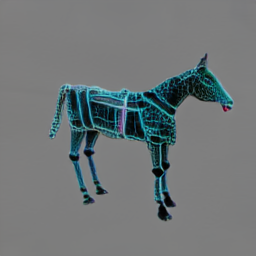} &
		\includegraphics[width=0.09\linewidth]{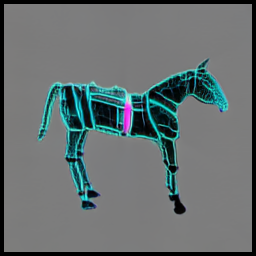} \\
	\end{tabular}
	\vspace{-0.5em}
	
	\begin{tabular}{*{3}c}
		\scriptsize{``a pig, cartoon style, exaggerated features"} & 
		\scriptsize{\scriptsize{\hspace{80pt}}``a duck, photorealistic, extremely detailed"\scriptsize{\hspace{60pt}}} & 
		\scriptsize{``a horse, neon lights, cybernetic implants"} \\
	\end{tabular}

	\begin{tabular}{*{11}c}
		\includegraphics[width=0.09\linewidth]{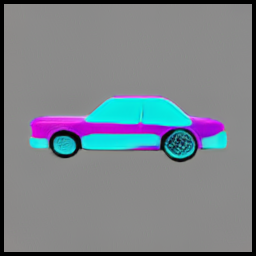} &
		\includegraphics[width=0.09\linewidth]{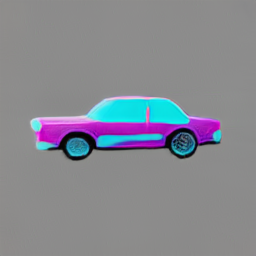} &
		\includegraphics[width=0.09\linewidth]{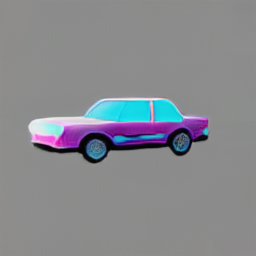} &
		\includegraphics[width=0.09\linewidth]{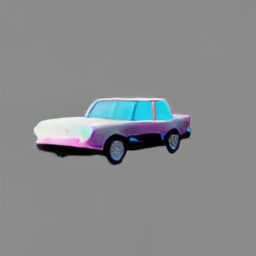} &
		\includegraphics[width=0.09\linewidth]{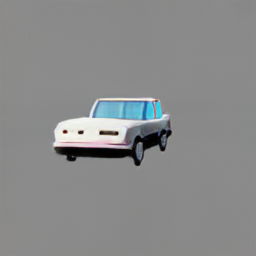} &
		\includegraphics[width=0.09\linewidth]{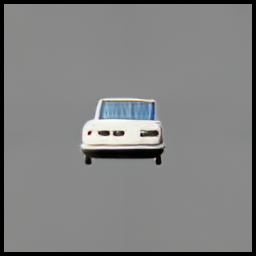} &
		\includegraphics[width=0.09\linewidth]{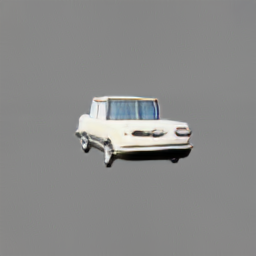} &
		\includegraphics[width=0.09\linewidth]{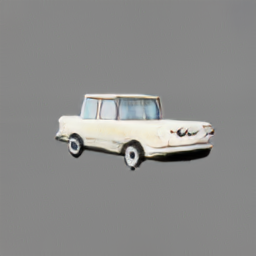} &
		\includegraphics[width=0.09\linewidth]{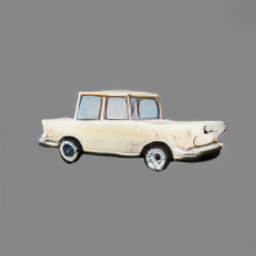} &
		\includegraphics[width=0.09\linewidth]{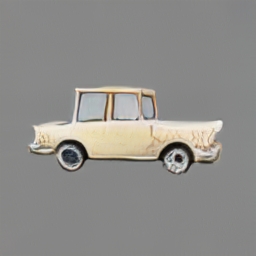} &
		\includegraphics[width=0.09\linewidth]{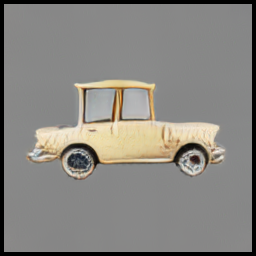} \\
	\end{tabular}
	\vspace{-0.5em}
	
	\begin{tabular}{*{3}c}
		\scriptsize{``a car, bright and bold colors, abstract"} & 
		\scriptsize{\scriptsize{\hspace{85pt}}``a car, documentary realism, unscripted"\scriptsize{\hspace{65pt}}} & 
		\scriptsize{``a car, primitive art, simple and symbolic"} \\
	\end{tabular}
	
	\vspace{-0.5em}
	\caption{ {\textbf{Text embedding interpolation results.}} The experimental results demonstrate that we can linearly interpolate between two text prompt embeddings. It shows the interpolation properties of GANs continue to be considerably smooth.}
	\label{fig_interpolation}
	\vspace{-1.5em}
\end{figure*}

\begin{figure*}
	\centering
	\begin{tabular}{c}
		
		\includegraphics[width=0.95\linewidth]{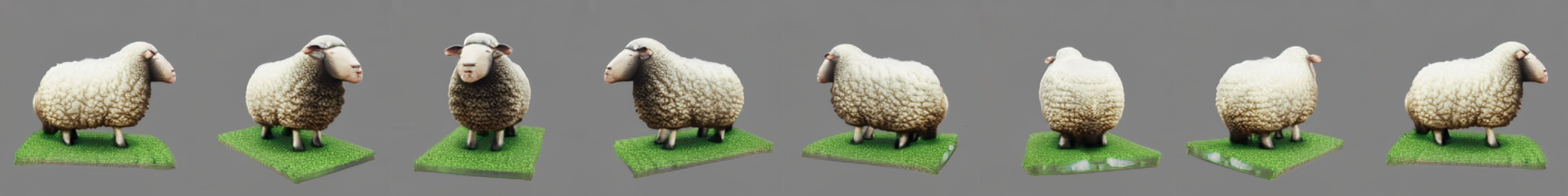} \\
		\small{``a sheep, sustainable art style, eco-friendly materials, environmental themes"} \\ 
		
		\includegraphics[width=0.95\linewidth]{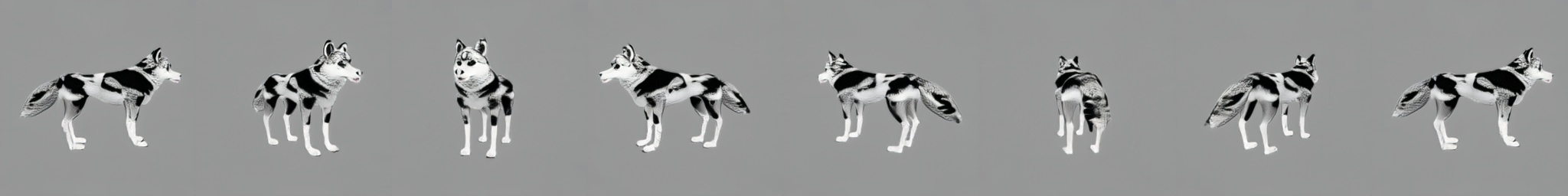} \\
		\small{``a wolf, manga style, Japanese comics, black and white"} \\ 
		
		
		\includegraphics[width=0.95\linewidth]{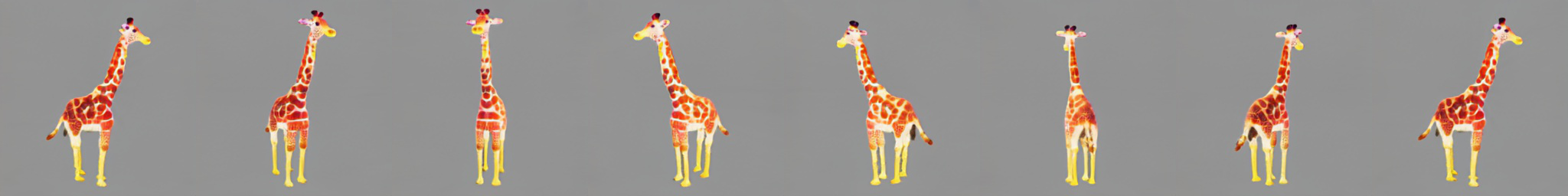} \\
		\small{``a giraffe, animate, high quality, cartoon style, exaggerated features, bright colors"} \\ 
		
		\includegraphics[width=0.95\linewidth]{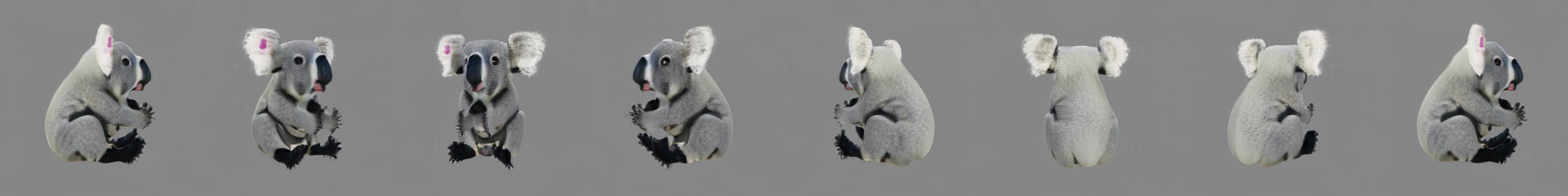} \\
		\small{``a koala, cinema verite style, documentary realism, unscripted"} \\ 
		
		\includegraphics[width=0.95\linewidth]{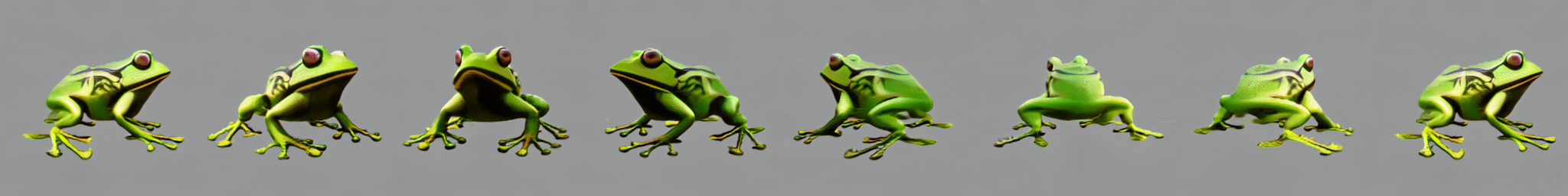} \\
		\small{``a frog, retro-futurist style, futuristic visions from the past, vintage aesthetics"} \\ 
		
		\includegraphics[width=0.95\linewidth]{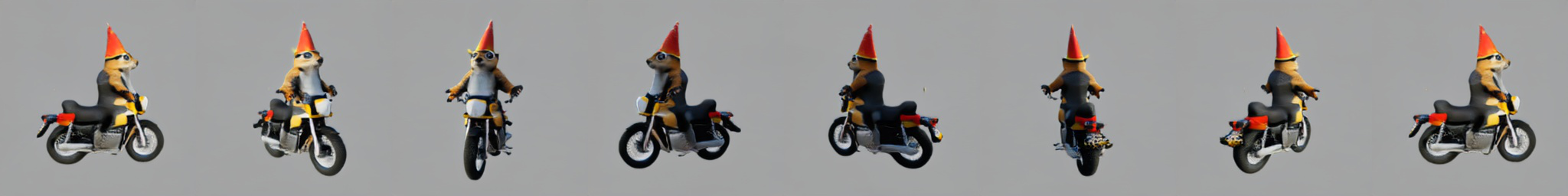} \\
		\small{``a squirrel, riding a motorcycle, wearing a leather jacket, wearing a party hat"} \\ 
		
		\includegraphics[width=0.95\linewidth]{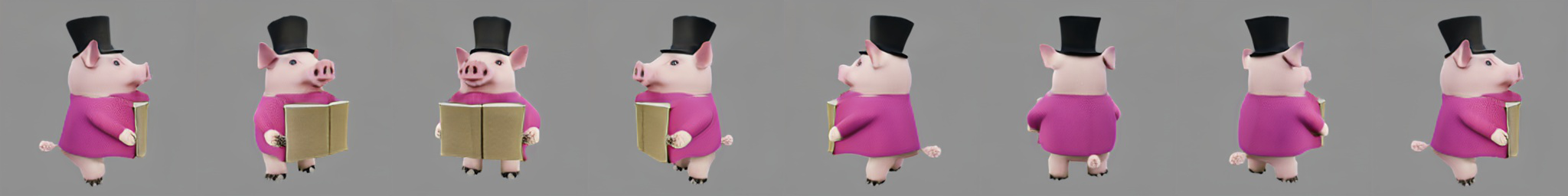} \\
		\small{``a pig, holding a book, wearing a sweater, wearing a tophat"}
		
	\end{tabular}
	\vspace{-0.9em}
	\caption{ {\textbf{Compositional generalization performance on unseen text prompts.}} It demonstrates that our network is able to generalize to unseen text prompts and generates high fidelity 3D assets.}
	\label{fig_generalize}
	\vspace{-1.4em}
\end{figure*}

\subsection{Qualitative comparisons}
The qualitative evaluation results are presented in \figref{fig_quality}. We present rendered images from two different views to evaluate their 3D consistency and texture quality. The surface normal or 3D mesh is also presented for geometry comparisons. The experimental results demonstrate that DreamFusion tends to generate over-saturated and blurry 3D assets due to the inherited characteristic of SDS loss. While ProlificDreamer improves the SDS loss and delivers impressive results, it still performs poorly for some text prompts. In contrary, our method delivers better results, in terms of both the texture quality and 3D consistency. It demonstrates the great potential to exploit multi-view images, generated by a pretrained image diffusion model, for high-quality text-to-3D content generation.

To demonstrate the generalization capability of our model to unseen prompts, we follow the experimental setting used by ATT3D \cite{lorraine2023att3d}. In particular, it generates compositional prompts using the template "a \{animal\} \{activity\} \{theme\}" and withholding a subset of prompts as unseen for evaluation. Additionally, we further select 40 animals and 40 styles, employing the compositional prompt "a \{animal\}, \{style\}" and exploit the compositions along the diagonal direction to validate style generalization. The 3D objects generated from part of these unseen prompts are illustrated in \ref{fig_generalize}. To better perceive consistent 3D objects, we render images of the same object from multi-views, \ie $0^{\circ}$, $90^{\circ}$, $180^{\circ}$ and $270^{\circ}$. Please refer to the Appendix and supplementary materials for more results. 
The experimental demonstrates that our network is able to generalize to unseen text prompts and delivers good results for both geometry and texture style compositions. 


\figref{fig_interpolation} presents the textual embedding interpolation results. We linearly interpolate the latent vector of two text prompts and input it to the generator network. It demonstrates that the interpolation properties of GANs continue to be considerably smooth.

%% file: sec/5_con.tex
\section{Conclusion and Future Work}
\label{sec:con}
\vspace{-0.5em}
We present a novel framework for efficient text-to-3D generation. Our network is built upon an unconditional 3D GAN network, and is trained via multi-view distillation of a pretrained text-to-multi-view model. 
Different from prior Score Distillation Sampling (SDS) based optimization methods, which usually requires large amount of computational resources to generate a 3D asset, we are able to generate a 3D object in only 8 ms once the network is trained.  
Due to the available resources, we currently only trained our network with a small amount of text prompts. Even with such limited number of text prompts, our network exhibit good generalization performance to unseen text input. It demonstrates the great potential of our framework for large-scale efficient text-to-3D generation task.  